\pdfoutput=1

\documentclass[11pt]{article}

\usepackage{acl2023}

\usepackage{times}
\usepackage{latexsym}

\usepackage[T1]{fontenc}

\usepackage[utf8]{inputenc}

\usepackage{microtype}

\usepackage{inconsolata}

\newcommand*\samethanks[1][\value{footnote}]{\footnotemark[#1]}
\usepackage{booktabs}
\usepackage{xcolor}
\usepackage{tabularx}
\usepackage{multirow}
\usepackage{amsmath}
\usepackage{amsfonts}
\usepackage{subcaption}
\usepackage{pgfplots}
\usepackage{pgfplotstable}
\usepackage{pifont}
\usepackage{colortbl}
\usepackage{CJK}
\usepackage{arydshln}
\newcommand{\abbr}{MP$^2$}

\title{Multitask Pre-training of Modular Prompt for Chinese Few-Shot Learning}

\author{
Tianxiang Sun\thanks{\ \ \ Equal contribution.}\quad\quad
Zhengfu He\samethanks\quad\quad
Qin Zhu\quad\quad
Xipeng Qiu\thanks{\ \ \ Corresponding author.}\quad\quad
Xuanjing Huang\\
School of Computer Science, Fudan University\\
Shanghai Key Laboratory of Intelligent Information Processing, Fudan University\\
\texttt{\{txsun19,zfhe19,xpqiu,xjhuang\}@fudan.edu.cn}\quad \quad
\texttt{zhuq22@m.fudan.edu.cn}
}

\begin{document}
\maketitle
\begin{abstract}
Prompt tuning is a parameter-efficient approach to adapting pre-trained language models to downstream tasks.
Although prompt tuning has been shown to match the performance of full model tuning when training data is sufficient, it tends to struggle in few-shot learning settings.
In this paper, we present \textbf{M}ulti-task \textbf{P}re-trained \textbf{M}odular \textbf{P}rompt (\textbf{\abbr}) to boost prompt tuning for few-shot learning.
\abbr~is a set of combinable prompts pre-trained on 38 Chinese tasks.
On downstream tasks, the pre-trained prompts are selectively activated and combined, leading to strong compositional generalization to unseen tasks.
To bridge the gap between pre-training and fine-tuning, we formulate upstream and downstream tasks into a unified machine reading comprehension task.
Extensive experiments under two learning paradigms, i.e., gradient descent and black-box tuning, show that \abbr~significantly outperforms prompt tuning, full model tuning, and prior prompt pre-training methods in few-shot settings.
In addition, we demonstrate that \abbr~can achieve surprisingly fast and strong adaptation to downstream tasks by merely learning 8 parameters to combine the pre-trained modular prompts.
\end{abstract}

\section{Introduction}
Pre-trained models (PTMs)~\citep{Devlin2019BERT,Lewis2020BART,Raffel2020T5,Qiu2020survey} with prompt-based learning have achieved remarkable progress in few-shot learning. A major reason behind their success is the closed gap between upstream pre-training and downstream fine-tuning~\cite{Liu2021PromptSurvey,Sun2021Paradigm}. Since the downstream tasks are reformulated into a unified (masked) language modeling ((M)LM for short) task, one can reuse the pre-trained (M)LM head instead of training a randomly initialized classification head to solve tasks with limited data. However, prompt-based learning (e.g., PET~\cite{Schick21PET} and LM-BFF~\cite{Gao20Making}) usually fine-tunes all the parameters of the PTM for each downstream task, which can be computationally expensive and deployment-inefficient, especially for large PTMs such as GPT-3~\cite{Brown2020GPT3}. 

Recently, much effort has been devoted to parameter-efficient prompt tuning~\citep{Li2021Prefix,Lester2021Prompt,Liu2021PTuning,Sun2022BBT}, which only learns a small number of soft prompt parameters while keeping the main body of the PTM untouched. In contrast to full model tuning, prompt tuning can get specialized models for specific tasks by simply attaching task-specific prompts, and therefore is highly efficient for serving different tasks. Though it has been demonstrated that prompt tuning can match the performance of full model tuning when training data is sufficient~\citep{Lester2021Prompt}, the soft prompt cannot be well trained from scratch in few-shot learning settings~\citep{Gu2021PPT} because the randomly initialized soft prompt introduces a new gap between pre-training and fine-tuning.

\begin{figure}[t!]
    \centering
    \includegraphics[width=\linewidth]{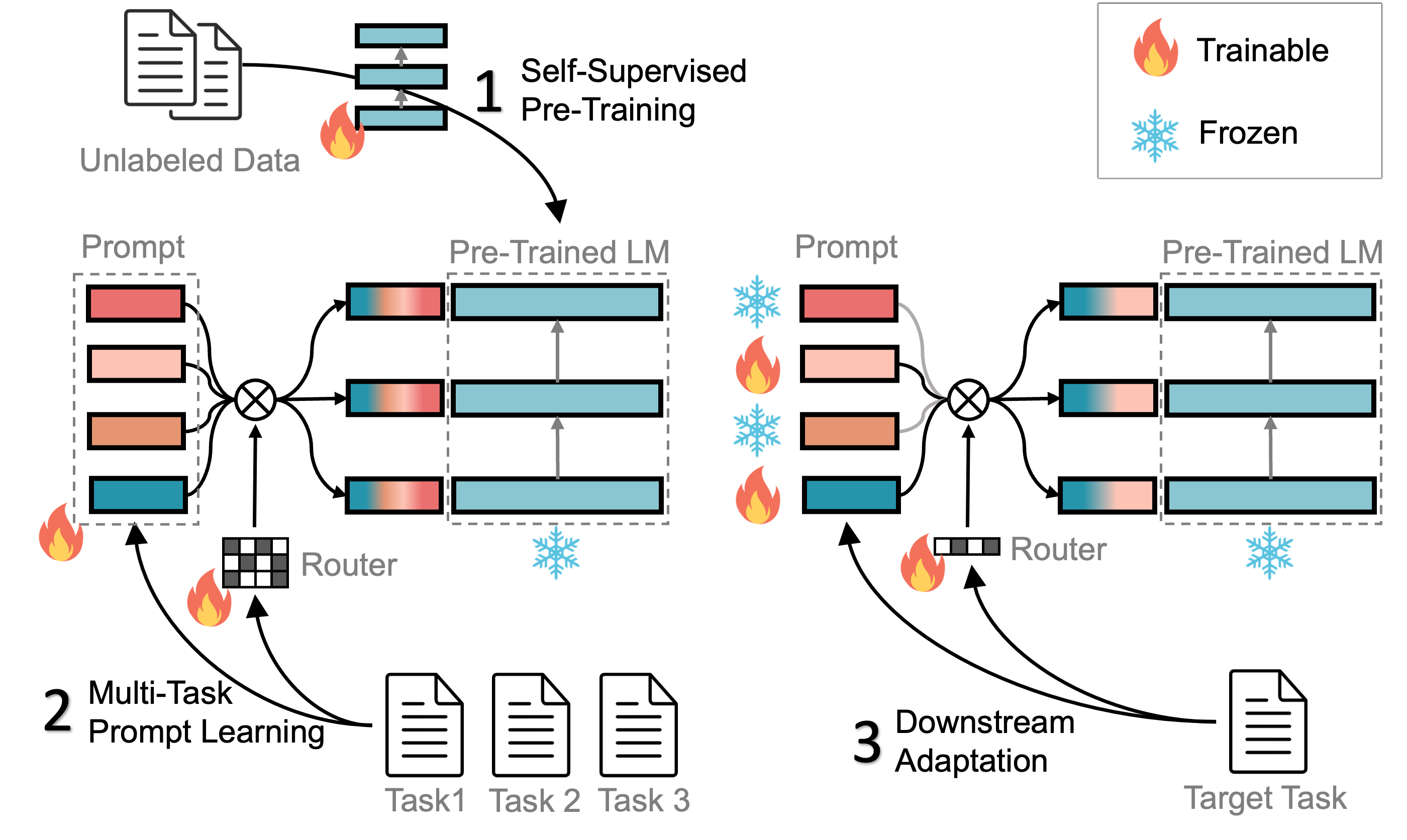}
    \caption{\abbr~achieves fast adaptation to downstream tasks through three steps: (1) Self-supervised pre-training on large-scale unlabeled data. (2) Pre-training modular prompts and the corresponding router with multi-task learning. (3) A subset of prompts is activated and tuned for adaptation to downstream tasks.}
    \label{fig:mpmp_paradigm}
\end{figure}

To bridge the gap between pre-training and fine-tuning for prompt tuning, we present \textbf{M}ulti-task \textbf{P}re-trained \textbf{M}odular \textbf{P}rompt (\textbf{\abbr}). As illustrated in Figure~\ref{fig:mpmp_paradigm}, we insert a second pre-training procedure before downstream fine-tuning, in which we pre-train a set of modular prompts with multi-task learning. The modular prompts are selectively activated and combined by a trainable router for specific tasks. By this, we can achieve fast adaptation to downstream tasks by learning to combine and reuse the pre-trained modular prompts. Drawing inspiration from the success of deep prompt tuning~\citep{Li2021Prefix,Liu2021PTuningv2}, we inject soft prompt into every layer of the PTM. Further, considering that a variety of tasks cannot be reformulated into a (M)LM task, we instead recast upstream and downstream tasks into a unified machine reading comprehension (MRC) task, which has shown great potential to unify various NLP tasks~\citep{McCann2018DecaNLP,Sun2021Paradigm}.

We pre-train \abbr~on 38 Chinese NLP tasks and evaluate on 14 downstream tasks including sentiment analysis, topic classification, natural language inference, question answering, multiple choice classification, and keyword extraction. Experimental results in few-shot learning settings demonstrate that \abbr~outperforms prompt tuning, full model tuning, and previous prompt pre-training methods~\cite{Gu2021PPT,Vu2022SPoT} by a large margin. We also evaluate the compatibility of \abbr~with black-box tuning (BBT)~\cite{Sun2022BBT} and BBTv2~\cite{Sun2022BBTv2}, which are gradient-free prompt tuning methods. As a result, \abbr~achieves significant improvement over BBT and BBTv2. Besides, we demonstrate that \abbr~can achieve surprisingly fast adaptation to target tasks by merely tuning the router (only 8 parameters) while freezing the PTM and all the prompts.\footnote{Code and data are publicly available at \url{https://github.com/Hzfinfdu/MPMP}.}

\section{Related Work}
This work lies in the line of parameter-efficient tuning (PET)~\cite{He2021Unified,Ding2022Delta}, which trains a small portion of parameters to adapt PTMs to downstream tasks. The small tunable parameters can be lightweight neural adapters between PTM layers~\cite{Houlsby2019Adapter}, or soft prompt attached to the input examples~\cite{Lester2021Prompt} or hidden states~\cite{Li2021Prefix}, or bias terms in the PTM parameters~\cite{Zaken22BitFit}, or low-rank matrices to be added to attention weights~\cite{Hu2021LoRA}. Especially, this work is closely related to two prior works on prompt tuning, namely PPT~\cite{Gu2021PPT} and SPoT~\cite{Vu2022SPoT}.

\begin{table}[t]
    \centering
    \resizebox{\linewidth}{!}{
    \begin{tabular}{lccr}
    \toprule
        \textbf{Method} & \textbf{Params.} & \textbf{Data Size} & \textbf{Data/Param.} \\ \midrule
        PPT & 410K & 10 GB & 24.39 GB/M \\
        \abbr~(Ours) & 307M & 15 GB & 0.05 GB/M \\
        \midrule
        BERT & 335M & 16 GB & 0.05 GB/M \\
        XLNet & 335M & 158 GB & 0.47 GB/M\\
        RoBERTa & 355M & 160 GB & 0.48 GB/M \\
        BART & 406M & 160 GB & 0.39 GB/M\\
        T5 & 11B & 745 GB & 0.07 GB/M\\
        \bottomrule
    \end{tabular}
    }
    \caption{Comparison of model size and data size for various pre-training methods. In contrast to conventional PTMs, there is a mismatch between the number of learnable parameters and the volume of training data for PPT.}
    \label{tab:model_vs_data}
\end{table}

\paragraph{Comparison with PPT.}
A prior work with the similar motivation is Pre-trained Prompt Tuning (PPT)~\citep{Gu2021PPT}, which pre-trains soft prompt prepended to the input embedding on large-scale unlabeled corpora with an objective of next sentence prediction (NSP). Different from the NSP in BERT~\citep{Devlin2019BERT}, PPT recasts the NSP task into a multiple choice classification (MCC) format. For downstream tasks, PPT formulates three types of tasks, namely single-sentence, sentence-pair, and multiple choice classification, into a unified MCC format such that the gap between the pre-training task and downstream tasks can be filled. Despite their success, we argue that PPT has three possible defects: \textbf{(1) Complexity Mismatch}: The number of learnable parameters and the volume of training data are mismatched. PPT trains 410K parameters with 10 GB training data. By contrast, conventional PTMs have much smaller data-parameter ratios (see Table~\ref{tab:model_vs_data}). Hence, the limited number of parameters can hardly contain the rich knowledge in the large corpora. \textbf{(2) Simple Objective}: The pre-training objective of PPT, i.e., NSP, is not difficult enough. It has been shown that the impact of the NSP objective is unreliable~\citep{Yang2019XLNet,Liu2019roberta}. As formulated by \citet{Lan2020ALBERT}, NSP can be accomplished through two subtasks, \textit{topic prediction} and \textit{coherence prediction}. Nevertheless, topic prediction is easier to learn than coherence prediction, and therefore can dominate learning and makes NSP a rather simple task. \textbf{(3) Limited Task}: The downstream tasks handled by PPT are limited. PPT cannot address tasks that cannot be reformulated into a MCC task, such as question answering. Besides, when pre-training with the MCC format, PPT supports up to 16 options (\texttt{A-P}), which means it only promises to adapt to tasks with no more than 16 labels. In this work, the above issues are well addressed by \abbr. \textbf{First}, \abbr~increases capacity of prompt in two dimensions, i.e., depth (deep prompt) and width (modular prompt), to match the complexity of training data. \textbf{Second}, \abbr~is pre-trained on 38 real-world Chinese tasks with multi-task learning, instead of pre-training in a self-supervised fashion with the NSP loss. \textbf{Third}, \abbr~recasts upstream and downstream tasks into a unified MRC task to support a wider range of downstream tasks.

\paragraph{Comparison with SPoT.}
Another work that is similar to ours is Soft Prompt Transfer (SPoT)~\cite{Vu2022SPoT}, which also explored training soft prompt with multi-task learning and then using it to initialize the prompt for a target task. By comparison, our proposed \abbr~has three main differences from SPoT: (1) We pre-train a set of modular prompts that are selectively combined and attached to every layer of the PTM rather than training a single prompt to be prepended merely to the input layer. (2) We formulate upstream and downstream tasks into a unified MRC task instead of unifying tasks into a text-to-text format~\cite{Raffel2020T5} where the output label words cannot be shared between upstream and downstream tasks.\footnote{A shared set of label words in prompt pre-training can be crucial to few-shot leaning. For example, PPT recasts tasks into the MCC format such that the label words are constrained to option words, i.e., \{\texttt{A}, \texttt{B}, \texttt{C}, \dots\}.} (3) Unlike SPoT that is mainly evaluated in full data settings, \abbr~is dedicated to few-shot learning.

\section{Methods}
We first introduce the MRC format used to unify different tasks in §\ref{sec:mrc}, and then describe the deep modular prompt in §\ref{sec:deep_mod_prompt}, and finally we detail the procedure of multi-task pre-training and downstream fine-tuning in §\ref{sec:pretrain} and §\ref{sec:finetune}, respectively.

\begin{figure*}[t!]
    \centering
    \includegraphics[width=.9\linewidth]{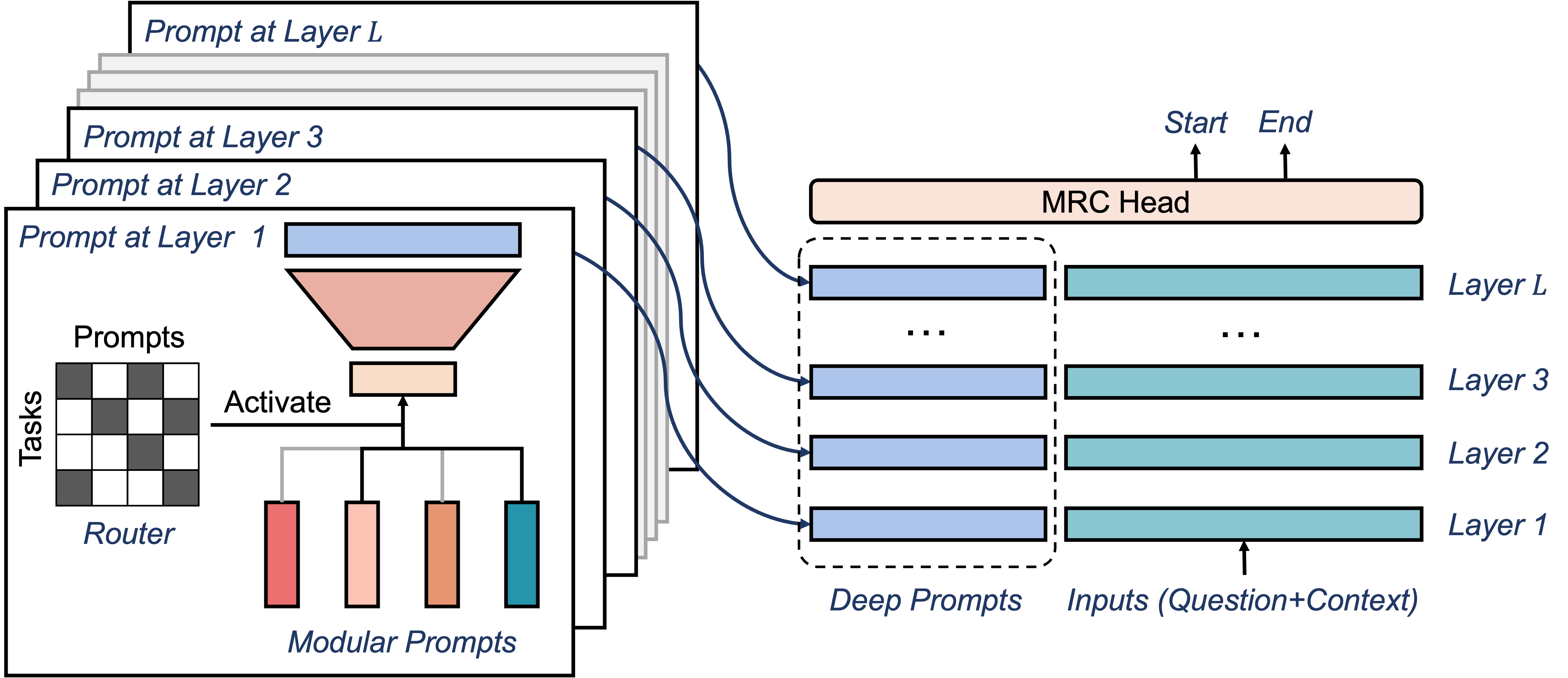}
    \caption{An illustration of the deep modular prompt during pre-training.}
    \label{fig:model_overview}
\end{figure*}

\subsection{Unifying Tasks with MRC}
\label{sec:mrc}
Bridging the gap between upstream and downstream tasks is crucial for few-shot learning. Prompt-based learning~\cite{Liu2021PromptSurvey} formulates downstream tasks into a (M)LM task, which, however, cannot cover a wide range of tasks. Besides, the label words (a.k.a. verbalizer) can be different across tasks. Therefore, the soft prompt pre-trained with a certain set of label words can be less effective to be used in a target task with a different set of label words.
To that end, PPT~\cite{Gu2021PPT} recasts upstream and downstream tasks into a MCC task such that different tasks can share the same set of label words, i.e., 16 option indicators (\texttt{A-P}). As a result, there is still a gap between pre-training and fine-tuning when performing classification with more than 16 labels. In addition, the task types supported by MCC can still be limited.

In \abbr, we adopt a more general format, machine reading comprehension (MRC), to unify upstream and downstream tasks. MRC has achieved great success in unifying a variety of NLP tasks~\cite{Sun2021Paradigm}. The input of MRC is comprised of a \textit{passage} (also referred to as \textit{context}) and a \textit{query}, and the output is the \textit{answer} of the query, which is a span of text in the input. Typically, the prediction of the answer is achieved by two binary classification heads on each token of the input, one for predicting the start position and one for predicting the end position~\cite{Xiong2017Dynamic,Seo2017BiDAF}.

For classification tasks, we use the original sample as the \textit{context} and construct a \textit{query} consisting of all possible labels. In contrast to PPT that pre-defines a set of option indicators, \abbr~ directly extracts the answer from the query, and therefore can generalize across tasks with different numbers of labels. Appendix~\ref{sec:append_format} contains some examples of converting tasks into the MRC format.

\subsection{Deep Modular Prompt}
\label{sec:deep_mod_prompt}
To increase the capacity of the soft prompt such that it can match the complexity of training data, we extend soft prompt in two dimensions, depth and width. Figure~\ref{fig:model_overview} provides an overview of the deep modular prompt.

\paragraph{Deep Prompt.}
Inspired by the success of deep prompt tuning~\cite{Li2021Prefix,Qin21Learning,Liu2021PTuningv2}, we inject soft prompt to every layer of the PTM instead of the mere input layer. The incorporation of deep prompt increases the number of learnable parameters and so as the adaptation ability to hard tasks.

\paragraph{Modular Prompt.}
For the soft prompt attached to each layer of the PTM, we extend the single static prompt to a set of modular prompts. Formally, we pre-train $K$ soft prompts $\{\mathbf{p}_1^{(l)}, \dots,\mathbf{p}_K^{(l)}\}$ for each layer $l$. For a certain task, the prompt at layer $l$ is the weighted mean of the set of soft prompts,
\begin{align}
    \mathbf{p}^{(l)} = \frac{1}{K}\sum_{k=1}^K w_k^{(l)}\mathbf{p}_k^{(l)},
\end{align}
where $\mathbf{w}^{(l)}=\{w_1^{(l)}, \dots, w_K^{(l)}\}$ are layer- and task-specific learnable parameters called \textit{router}. To pursue compositional generalization, we encourage the prompts to be sparsely activated and combined. Thus, the router $\mathbf{w}^{(l)}$ should be binary-valued, i.e., $\mathbf{w}^{(l)}\in\{0,1\}^{K}$. Each single prompt can be viewed as some fundamental skill, and a task can be solved by combining such modular skills. Different tasks tend to require different subsets of the skills. Though similar ideas have been proposed in other names and contexts~\cite{Sun2020Sparse,zhang2022skillnetnlu,Ponti2022Modular}, this is the first work that implements the skills with soft prompts to drive pre-trained language models.

\paragraph{Relaxed Bernoulli Distribution.}
A challenge is that the discrete router $\mathbf{w}$~\footnote{For simplicity, we omit the superscript $(l)$ without causing confusion.} is not differentiable and therefore cannot be optimized by gradient descent in an end-to-end fashion. To that end, we keep $\mathbf{w}\in\mathbb{R}^K$ as free parameters to parameterize a relaxed Bernoulli (or binary concrete) distribution~\cite{Maddison2017Concrete}, which can be considered as a continuous relaxation of the Bernoulli distribution. From the relaxed Bernoulli distribution, we sample $\mathbf{\hat{w}}$ to weight the modular prompts, i.e., $\mathbf{p} = \sum_{k=1}^K \hat{w}_k\mathbf{p}_k/K$. By using the reparameterization trick~\cite{Kingma2014VAE}, the router can be learned via gradient descent while maintaining some degree of stochasticity. Formally, the sampling procedure for $\hat{w}_k\sim \text{\texttt{RelaxedBernoulli}}(\alpha, \tau)$ is as follows,
\begin{align}
    u&\sim \text{\texttt{Uniform}}(0,1),\\
    v&=\log(\alpha)+\log(u)-\log(1-u),\\
    \hat{w}_k&=\sigma(v/\tau),
\end{align}
where $\alpha\in(0,\infty)$ is the location parameter, $\sigma$ is the Sigmoid function, and $\tau\in(0,\infty)$ is the temperature to control the degree of approximation. Note that $w_k$ can be negative during training and therefore cannot be used directly as the location parameter $\alpha$. To ensure that $\alpha\in (0,\infty)$, we set $\alpha$ as follows,
\begin{align}
    \alpha=\frac{\sigma(w_k)}{1-\sigma(w_k)}.
\end{align}

During inference, we simply set $\hat{w}_k=1$ if $w_k>0$, otherwise $\hat{w}_k=0$.

\paragraph{Intrinsic Reparameterization.}
Recent studies~\cite{Sun2022BBT,Diao2022Black} have demonstrated that prompt tuning can be achieved in a much lower dimensional \textit{intrinsic subspace} through gradient-free optimization.
To benefit tuning in the intrinsic subspace, we perform \textit{intrinsic reparameterization}, which is to decompose the original modular prompt $\mathbf{p}_k\in\mathbb{R}^D$ into an intrinsic prompt $\mathbf{z}_k\in\mathbb{R}^d$ and a projection matrix $\mathbf{A}\in\mathbb{R}^{D\times d}$.
Note that $\mathbf{A}$ is shared by the modular prompts $\{\mathbf{p}_k\}_{k=1}^K$ at the same layer.
During multi-task pre-training, both $\mathbf{z}_k$ and $\mathbf{A}$ are updated. On downstream tasks, black-box tuning (BBT)~\cite{Sun2022BBT} can be enabled by only tuning the intrinsic prompt $\mathbf{z}_k$ while keeping $\mathbf{A}$ frozen.

\subsection{Multi-Task Pre-Training}
\label{sec:pretrain}
Multi-task learning has been shown to boost the performance of prompt tuning in a variety of tasks~\cite{Vu2022SPoT}. Following their success, we pre-train the deep modular prompts on a mixture of 38 Chinese NLP tasks with varying types, domains, and sizes. To handle the unbalanced data sizes, for each forward computation, we first randomly sample a task ID from 1 to 38 and then fetch a batch of training data corresponding to the sampled task, such that the number of learning steps for each task is expected to be identical.

\paragraph{Fast and Slow Learning.}
For the pre-training of the routers and the prompts, we intuitively encourage fast learning for the routers to reuse existing modular prompts to adapt to the current task, and slow learning for the task-specific prompts. In particular, we adopt a higher learning rate for the routers $\mathbf{z}$ to change quickly, and adopt a lower learning rate for the modular prompts $\mathbf{p}$ to change slowly and stably. Similar ideas are also explored by \citet{Madan2021Fast,Ponti2022Modular}.

\subsection{Downstream Fine-Tuning}
\label{sec:finetune}
For fast adaptation to downstream tasks, we propose the \textit{two-stage tuning}. \textbf{In stage I}, we allocate a random router for each layer to a new target task and train the routers to selectively reuse pre-trained modular prompts to solve the target task while keeping all other parameters frozen. \textbf{In stage II}, we freeze the routers and only tune the selected prompts. The PTM parameters are unchanged throughout the entire fine-tuning process.

We explore fine-tuning \abbr~under two learning paradigms, namely \textit{gradient descent} and \textit{black-box tuning}. For gradient descent, we use an Adam~\cite{kingma2015adam} optimizer to perform two-stage tuning. For black-box tuning, we adopt the Bayesian optimization (BO)~\cite{Mockus1974Bayesian} in stage I to optimize the routers, and adopt the CMA-ES~\cite{Hansen2001CMA} to optimize the selected intrinsic prompts $\mathbf{z}_k$ while freezing the projection matrices $\mathbf{A}$. See Appendix~\ref{sec:append_implement} for detailed description of fine-tuning.

\section{Experiments}
\subsection{Datasets and Tasks}
\label{sec:data_task}
\paragraph{Pre-training Tasks.}
We collect 38 public Chinese NLP tasks ranging from different task types, domains, and data sizes as upstream tasks for pre-training.
The total size of the pre-training data is 15GB.
Appendix~\ref{sec:append_pretrain_task} contains full details of the pre-training tasks.

\paragraph{Downstream Tasks.}
We divide 14 downstream tasks into two tracks: \textsc{Unseen Data} and \textsc{Unseen Task}. The 7 tasks in the \textsc{Unseen Data} track are a subset of upstream tasks, for which we retain a small portion of training data from the pre-training corpora to ensure that the downstream samples are unseen to \abbr. The \textsc{Unseen Task} track is comprised of 7 tasks that are completely held-out tasks. Table~\ref{tab:downstream_task} contains statistics of the downstream tasks. The sources of the tasks are in Appendix~\ref{sec:append_pretrain_task}.

\begin{table}[t]
    \centering
    \resizebox{\linewidth}{!}{
\begin{tabular}{llccc}
\toprule
\textbf{Setting}                                                       & \textbf{Dataset} & \textbf{Task} & \textbf{|Test|} & \textbf{|Labels|} \\ \midrule
\multirow{7}{*}{\begin{tabular}[c]{@{}l@{}}\textsc{Unseen}\\ \textsc{Data}\end{tabular}} & Amazon           & TC            & 5789        &   5            \\
                                                                       & THUCNews         & TC            &      5000           &         10         \\
                                                                       & BQ               & NLI           & 10000            &   2        \\
                                                                       & CMNLI            & NLI           &             12545   &      3             \\
                                                                       & CMRC-2018        & MRC           &            2886     &       N/A            \\
                                                                       & CCPM             & MCQA          &              2720   &   4            \\
                                                                       & COTE-MFW         & KE            &           8251     &     N/A              \\ \midrule
\multirow{7}{*}{\begin{tabular}[c]{@{}l@{}}\textsc{Unseen}\\\textsc{Task}\end{tabular}} & ChnSent          & TC            &      1200           &           2       \\
                                                                       & TNews            & TC            & 10000           &         15          \\
                                                                       & OCNLI            & NLI           &  2950               &      3             \\
                                                                       & LCQMC            & NLI           &   8802             &       2            \\
                                                                       & DRCD             & MRC           &   1238             &        N/A           \\
                                                                       & C$^3$               & MCQA          &    1991             &       [2, 4]           \\
                                                                       & COTE-BD          & KE            &     1706            &     N/A              \\ \bottomrule
\end{tabular}
}
    \caption{Statistics of downstream tasks. TC: text classification. NLI: natural language inference. MRC: machine reading comprehension. MCQA: multiple choice question answering. KE: keyword extraction.}
    \label{tab:downstream_task}
\end{table}

\paragraph{True Few-Shot Setting.}
For downstream tasks, we follow the same procedure as \citet{Gu2021PPT} to form the true few-shot learning settings~\cite{Perez2021TrueFewShot}. 
In particular, we randomly draw 32 samples from the original training set to construct a few-shot training set $\mathcal{D}_{\text{train}}$, and construct a development set $\mathcal{D}_{\text{dev}}$ by randomly selecting another 32 samples from the original training set. We ensure that the number of labels is balanced for both training and development set. For tasks with more than 5 labels, we randomly select 8 samples for each label. We use the original development sets as the test sets. For datasets without development sets, we use the original test sets.  

\begin{table*}[t!]
\resizebox{\linewidth}{!}{
\begin{tabular}{lllccccccccc}
\toprule
\multicolumn{12}{c}{\textsc{Unseen Data}}\\ \midrule
\multirow{2}{*}{\textbf{Paradigm}} & \multirow{2}{*}{\textbf{Backbone}} & \multirow{2}{*}{\textbf{Methods}} & \textbf{Tunable} & \textbf{Amazon}  & \textbf{THUCNews} & \textbf{BQ}    & \textbf{CMNLI} & \textbf{CMRC-2018} & \textbf{CCPM} & \textbf{COTE-MFW} & \multirow{2}{*}{\textbf{Avg.}} \\
& & & \textbf{Params} & Acc. & Acc. & Acc. & Acc. & F1 & Acc. & F1 & \\ \midrule
\multirow{6}{*}{\begin{tabular}[c]{@{}l@{}}Gradient\\Descent\end{tabular}} & \multirow{3}{*}{\begin{tabular}[c]{@{}c@{}}CPM-2\\(11B)\end{tabular}} & Model Tuning & 11B & 42.5 \textsubscript{2.0} & - & - & 40.7 \textsubscript{1.0} & - & 81.8 \textsubscript{1.6} & - & - \\
& & Prompt Tuning & 410K & 30.3 \textsubscript{4.8} & - & - & 35.4 \textsubscript{0.5} & - & 31.0 \textsubscript{9.7} & - & - \\
& & PPT & 410K & 44.6 \textsubscript{1.1} & - & - & 40.6 \textsubscript{0.4} & - & 83.4 \textsubscript{0.9} & - & - \\ \cmidrule{2-12}
& \multirow{7}{*}{\begin{tabular}[c]{@{}c@{}}CPT\\(393M)\end{tabular}} & Model Tuning & 393M & 47.3 \textsubscript{5.3} & 93.5 \textsubscript{0.3} & 57.3 \textsubscript{1.7} & 34.7 \textsubscript{0.1} & 37.5 \textsubscript{7.4} & 76.1 \textsubscript{2.4} & 81.7 \textsubscript{1.3} & 61.2 \\
& & Prompt Tuning & 50K & 32.9 \textsubscript{2.4} & 68.6 \textsubscript{4.2} & 51.3 \textsubscript{0.7} & 33.8 \textsubscript{0.4} & 3.5 \textsubscript{0.4} & 27.3 \textsubscript{1.9} & 57.7 \textsubscript{1.0} & 39.3 \\
& & P-Tuning v2 & 1.2M & 47.7 \textsubscript{2.3} & 90.4 \textsubscript{0.6} & 54.6 \textsubscript{1.6} & 34.5 \textsubscript{0.2} & 34.4 \textsubscript{10.4} & 76.3 \textsubscript{2.0} & 81.8 \textsubscript{2.0} & 60.0 \\ 
& & PPT & 50K & 49.7 \textsubscript{2.3} & 87.9 \textsubscript{1.3} & 53.3 \textsubscript{0.9} & 34.2 \textsubscript{0.6} & 6.1 \textsubscript{0.6} & 83.1 \textsubscript{1.2} & 74.0 \textsubscript{4.1} & 55.5 \\
& & SPoT & 50K & 55.2 \textsubscript{2.9} & 89.4 \textsubscript{0.9} & 61.1 \textsubscript{1.5} & 39.0 \textsubscript{0.5} & 56.6 \textsubscript{1.7} & 85.2 \textsubscript{0.5} & 86.5 \textsubscript{0.7} & 67.6 \\ \cmidrule{3-12} 
& & Shallow \abbr & 50K$\sim$400K & 62.3 \textsubscript{1.0} & 91.2 \textsubscript{1.6} & 71.8 \textsubscript{2.0} & 66.5 \textsubscript{2.3} & 68.6 \textsubscript{2.3} & 85.3 \textsubscript{1.8} & 87.4 \textsubscript{1.2} & 76.2 \\
& & Deep \abbr & 1.2M$\sim$9.6M & \textbf{65.3} \textsubscript{1.7} & \textbf{95.2} \textsubscript{0.2} & \textbf{81.4} \textsubscript{1.3} & \textbf{76.3} \textsubscript{0.8} & \textbf{82.8} \textsubscript{1.0} & \textbf{92.4} \textsubscript{0.3} & \textbf{90.5} \textsubscript{0.2} & \textbf{83.4} \\ \midrule
\multirow{6}{*}{\begin{tabular}[c]{@{}l@{}}Black-Box\\ Tuning\end{tabular}} & \multirow{6}{*}{\begin{tabular}[c]{@{}c@{}}CPT\\(393M)\end{tabular}} & BBT & 300 & 44.5 \textsubscript{1.5} & 49.2 \textsubscript{6.0} & 51.7 \textsubscript{0.5} & 35.4 \textsubscript{0.7} & - & 26.4 \textsubscript{0.5} & - & - \\
& & BBTv2 & 7.2K & 47.7 \textsubscript{1.7} & 84.0 \textsubscript{0.8} & 53.5 \textsubscript{0.8} & 37.8 \textsubscript{0.4} & - & 26.9 \textsubscript{1.5} & - & - \\ \cmidrule{3-12} 
& & Shallow \abbr & 308 & 58.5 \textsubscript{5.1} & 92.4 \textsubscript{0.4} & 75.2 \textsubscript{0.8} & 66.4 \textsubscript{1.4} & 75.6 \textsubscript{1.9} & 90.6 \textsubscript{0.2} & 88.1 \textsubscript{1.1} & 78.1\\
& & \ - Router-only & 8 & 62.5 \textsubscript{3.9} & 92.6 \textsubscript{0.5} & 75.6 \textsubscript{0.8} & 63.4 \textsubscript{3.3} & 77.7 \textsubscript{0.6} & 90.3 \textsubscript{0.7} & 89.2 \textsubscript{0.6} & 78.7\\
& & Deep \abbr & 7.4K & 66.0 \textsubscript{1.0} & \textbf{94.6} \textsubscript{0.2} & \textbf{80.9} \textsubscript{0.8} & \textbf{76.3} \textsubscript{2.1} & 83.9 \textsubscript{0.8} & \textbf{92.4} \textsubscript{0.7} & 90.1 \textsubscript{0.2} & \textbf{83.5} \\ 
& & \ - Router-only & 192 & \textbf{66.1} \textsubscript{0.5} & \textbf{94.6} \textsubscript{0.2} & \textbf{80.9} \textsubscript{0.8} & 74.2 \textsubscript{2.2} & \textbf{84.0} \textsubscript{0.9} & 91.8 \textsubscript{0.7} & \textbf{90.2} \textsubscript{0.2} & 83.1\\
\midrule
\multicolumn{12}{c}{\textsc{Unseen Task}}\\ \midrule
\multirow{2}{*}{\textbf{Paradigm}} & \multirow{2}{*}{\textbf{Backbone}} & \multirow{2}{*}{\textbf{Methods}} & \textbf{Tunable} & \textbf{ChnSent} & \textbf{TNews}    & \textbf{OCNLI} & \textbf{LCQMC} & \textbf{DRCD}      & \textbf{C$^3$}   & \textbf{COTE-BD}  & \multirow{2}{*}{\textbf{Avg.}} \\
& & & \textbf{Params} & Acc. & Acc. & Acc. & Acc. & F1 & Acc. & F1 & \\ \midrule
\multirow{6}{*}{\begin{tabular}[c]{@{}l@{}}Gradient\\Descent\end{tabular}} & \multirow{3}{*}{\begin{tabular}[c]{@{}c@{}}CPM-2\\(11B)\end{tabular}} & Model Tuning & 11B & 86.1 \textsubscript{1.8} & - & 38.5 \textsubscript{1.5} & 58.8 \textsubscript{1.8} & - & 38.4 \textsubscript{3.7} & - & - \\
& & Prompt Tuning & 410K & 62.1 \textsubscript{3.1} & - & 37.0 \textsubscript{0.5} & 51.5 \textsubscript{3.4} & - & 28.2 \textsubscript{0.4} & - & - \\
& & PPT & 410K & 90.7 \textsubscript{0.2} & - & 41.5 \textsubscript{1.5} & 55.0 \textsubscript{0.4} & - & 50.2 \textsubscript{0.6} & - & - \\ \cmidrule{2-12}
& \multirow{7}{*}{\begin{tabular}[c]{@{}c@{}}CPT\\(393M)\end{tabular}} & Model Tuning & 393M & 76.8 \textsubscript{2.9} & 47.8 \textsubscript{0.8} & 35.6 \textsubscript{1.6} & 55.3 \textsubscript{2.1} & 29.0 \textsubscript{9.7} & 30.0 \textsubscript{2.5} & 85.2 \textsubscript{1.4} & 51.4 \\ 
& & Prompt Tuning & 50K & 60.6 \textsubscript{2.9} & 27.0 \textsubscript{0.9} & 33.0 \textsubscript{1.8} & 49.2 \textsubscript{2.6} & 2.9 \textsubscript{0.2} & 25.5 \textsubscript{0.8} & 61.9 \textsubscript{1.2} & 37.2 \\ 
& & P-Tuning v2 & 1.2M & 75.9 \textsubscript{2.3} & 46.9 \textsubscript{0.8} & 33.7 \textsubscript{0.2} & 49.7 \textsubscript{2.2} & 17.8 \textsubscript{7.9} & 28.0 \textsubscript{3.7} & 86.2 \textsubscript{2.1} & 48.3\\ 
& & PPT & 50K & 64.1 \textsubscript{3.4} & 44.8 \textsubscript{0.9} & 34.2 \textsubscript{0.7} & 51.4 \textsubscript{2.1} & 5.0 \textsubscript{1.4} & 36.8 \textsubscript{2.4} & 77.5 \textsubscript{1.0} & 44.8 \\ 
& & SPoT & 50K & 87.0 \textsubscript{0.9} & 48.2 \textsubscript{1.2} & 38.7 \textsubscript{1.0} & 60.9 \textsubscript{2.1} & 57.8 \textsubscript{1.2} & \textbf{44.9} \textsubscript{0.8} & 88.1 \textsubscript{0.3} & 60.8 \\ \cmidrule{3-12} 
& & Shallow \abbr & 50K$\sim$400K & 90.5 \textsubscript{0.2} & 51.4 \textsubscript{1.1} & 53.4 \textsubscript{5.0} & 72.5 \textsubscript{1.9} & 67.2 \textsubscript{3.0} & 44.1 \textsubscript{1.6} & 88.8 \textsubscript{0.7} & 66.8 \\
& & Deep \abbr & 1.2M$\sim$9.6M & \textbf{92.0} \textsubscript{0.1} & \textbf{54.7} \textsubscript{0.3} & \textbf{64.1} \textsubscript{2.3} & \textbf{83.5} \textsubscript{1.0} & \textbf{80.6} \textsubscript{0.9} & 35.4 \textsubscript{0.9} & \textbf{91.8} \textsubscript{0.3} & \textbf{71.7} \\ \midrule
\multirow{6}{*}{\begin{tabular}[c]{@{}l@{}}Black-Box\\ Tuning\end{tabular}} & \multirow{6}{*}{\begin{tabular}[c]{@{}c@{}}CPT\\(393M)\end{tabular}} & BBT & 300 & 84.7 \textsubscript{1.7} & 35.5 \textsubscript{1.7} & 32.6 \textsubscript{0.4} & 50.7 \textsubscript{4.0} & - & 28.7 \textsubscript{1.1} & - & - \\
& & BBTv2 & 7.2K & 85.8 \textsubscript{0.8} & 47.2 \textsubscript{1.2} & 36.0 \textsubscript{1.0} & 56.6 \textsubscript{2.2} & - & 29.3 \textsubscript{0.4} & - & - \\ \cmidrule{3-12} 
& & Shallow \abbr & 308 & 90.2 \textsubscript{0.4} & 52.4 \textsubscript{1.0} & 54.0 \textsubscript{2.7} & 77.1 \textsubscript{1.8} & 73.4 \textsubscript{1.1} & 42.7 \textsubscript{0.9} & 89.7 \textsubscript{0.4} & 68.6 \\
& & \ - Router-only & 8 & 90.4 \textsubscript{0.3} & 49.9 \textsubscript{3.2} & 53.3 \textsubscript{3.8} & 72.6 \textsubscript{0.9} & 71.5 \textsubscript{0.8} & \textbf{43.7} \textsubscript{2.1} & 88.3 \textsubscript{0.9} & 67.1 \\
& & Deep \abbr & 7.4K & \textbf{91.7} \textsubscript{0.4} & \textbf{55.1} \textsubscript{0.9} & \textbf{65.7} \textsubscript{1.9} & \textbf{84.6} \textsubscript{0.9} & 79.2 \textsubscript{0.8} & 36.0 \textsubscript{0.5} & 91.5 \textsubscript{0.2} & \textbf{72.0}\\ 
& & \ - Router-only & 192 & \textbf{91.7} \textsubscript{0.4} & 54.3 \textsubscript{0.7} & 65.3 \textsubscript{2.3} & 83.4 \textsubscript{1.7} & \textbf{79.8} \textsubscript{1.2} & 36.8 \textsubscript{0.9} & \textbf{91.6} \textsubscript{0.1} & 71.8 \\
\bottomrule
\end{tabular}
}
\caption{Main results on downstream tasks. Results on CPM-2 are taken from \citet{Gu2021PPT} since our experimental settings are consistent. For PPT on CPM-2, we take the results of the "Unified PPT" reported in the original paper.}
\label{tab:main_results}
\end{table*}

\subsection{Backbones and Baselines}
We choose CPT-large~\cite{Shao2021CPT} as our backbone model, which is a competitive Chinese PTM consisting of a 20-layered shared encoder, a 4-layered understanding decoder and a 4-layered generation decoder. In our experiment, we use the encoder and the understanding decoder to compose a 24-layered PTM. We attach soft prompt to the input layer and all intermediate layers except the last layer, which has no effect on the output. Therefore, we pre-trained 24 sets of modular prompts, each corresponding to one layer of CPT. In addition to the pre-trained \textbf{Deep \abbr}, we also pre-trained a set of modular prompts that are merely attached to the input layer, denoted as \textbf{Shallow \abbr}.

We evaluate \abbr~under two learning paradigms: \textit{gradient descent} and \textit{black-box tuning}. For gradient descent, we consider \textbf{(1) Model Tuning}, which fine-tunes all parameters of the PTM; \textbf{(2) Prompt Tuning}~\cite{Lester2021Prompt}, which prepends a sequence of soft prompt tokens to the input and only tunes the soft prompt for adaptation; \textbf{(3) P-Tuning v2}~\cite{Liu2021PTuningv2}: which incorporates and tunes soft prompt at every layer of the PTM. Prompt tuning and p-tuning v2 can be seen as the baselines to Shallow \abbr~and Deep \abbr, respectively. Besides, we compare with two previous prompt pre-training methods: \textbf{(4) PPT}~\cite{Gu2021PPT}, which pre-trains soft prompt on large-scale unlabeled data with self-supervised learning; and \textbf{(5) SPoT}~\cite{Vu2022SPoT}, which pre-trains soft prompt with multi-task learning. For fair comparison, we reimplement PPT and SPoT with the same backbone model, i.e., CPT-large. For PPT, we pre-trained the "Unified PPT" on the same pre-training corpora as in the original paper, i.e., 10GB WuDaoCorpora~\cite{Yuan2021Wudaocorpora}. For SPoT, we pre-trained a single soft prompt with the same 38 Chinese NLP tasks as used by \abbr. Therefore, experiments of SPoT can be seen as an ablation study on the effect of the modular prompt. For black-box tuning, we consider two baselines: \textbf{(1) BBT}~\cite{Sun2022BBT}, which adopts a gradient-free optimizer to tune a low-dimensional intrinsic prompt, and then randomly embeds it into the original prompt space to be concatenated with the input embedding; and \textbf{(2) BBTv2}~\cite{Sun2022BBTv2}, which extends BBT by incorporating soft prompt into every layer of the PTM and uses a divide-and-conquer algorithm to alternately optimize the soft prompt at each layer. 

The prompt length is set to 50 for both shallow \abbr~and deep \abbr. Each set of modular prompts is consisting of $K=8$ soft prompts, and therefore the pre-trained routers are in the shape of $38\times8$. Shallow \abbr~has only one router while deep \abbr~contains 24 routers corresponding to 24 layers. Hyper-parameters and more implementation details are provided in Appendix~\ref{sec:append_implement}.

\begin{figure}[t]
    \centering
    \begin{subfigure}{\linewidth}
    \centering
    \includegraphics[width=\linewidth]{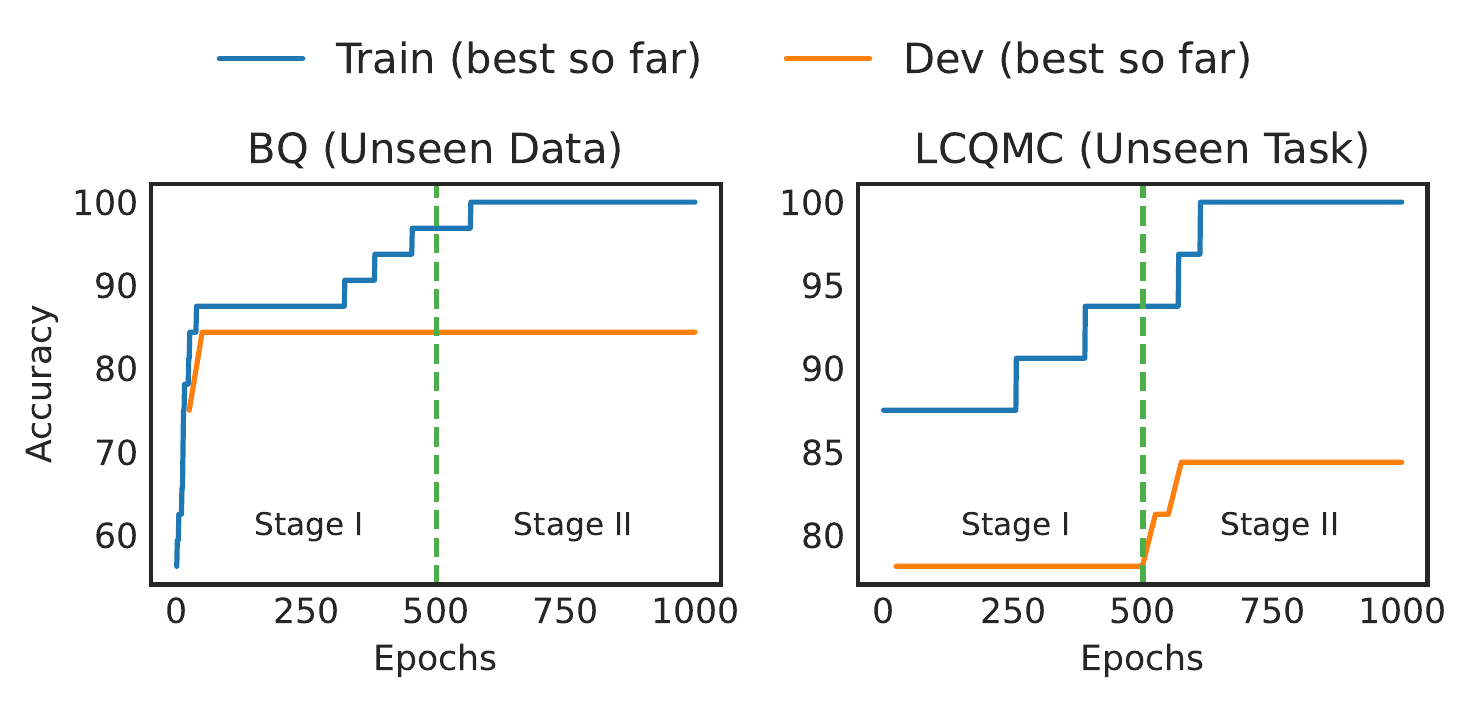}
    \caption{Gradient descent}
    \end{subfigure}
    \begin{subfigure}{\linewidth}
    \centering
    \includegraphics[width=\linewidth]{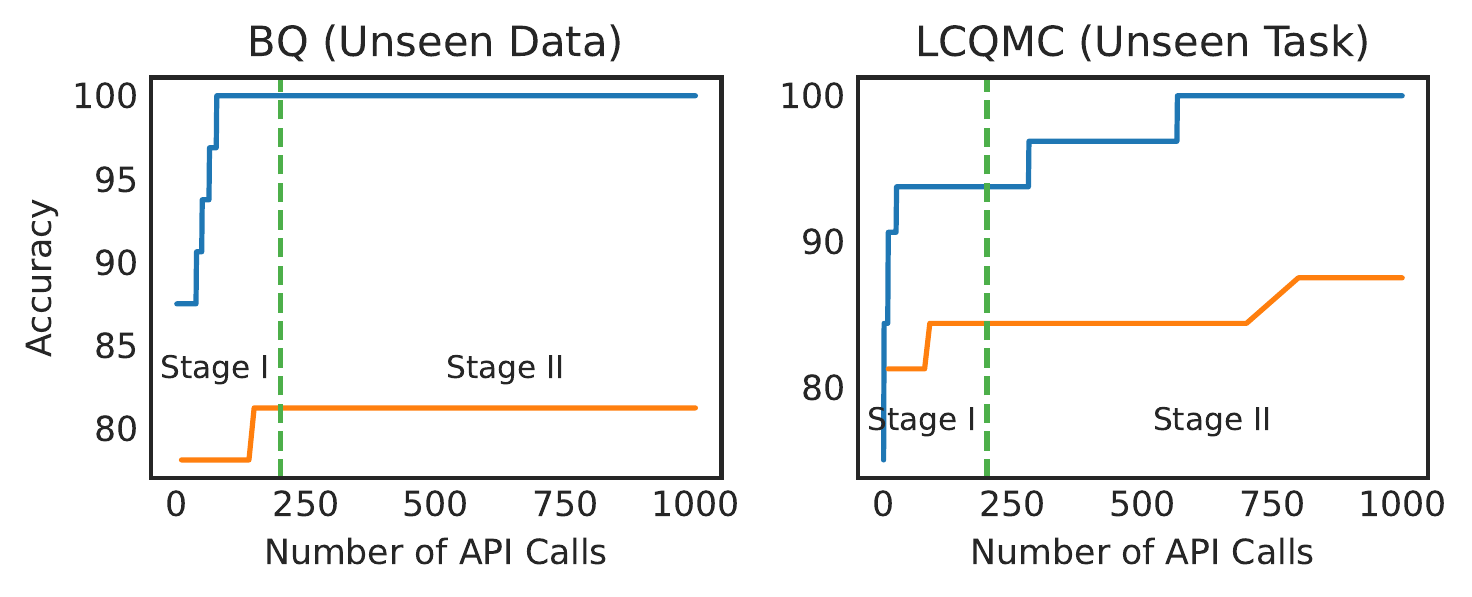}
    \caption{Black-box tuning}
    \end{subfigure}
    \caption{Two-stage tuning with shallow \abbr~for initialization under two learning paradigms. The green dashed lines indicate the boundary between the two stages.}
    \label{fig:two_stage}
\end{figure}

\subsection{Results}
\paragraph{Main Results.}
Main results on 14 downstream tasks are listed in Table~\ref{tab:main_results}.
We report mean and standard deviation of performance over 5 runs with different random seeds.
Overall, \abbr~outperforms all baselines by a large margin. By further comparison, we have the following findings: \textit{(1) Deep Prompt vs. Shallow Prompt}: Deep prompt methods (i.e., p-tuning v2, BBTv2, and deep \abbr) significantly outperform their corresponding shallow versions (i.e., prompt tuning, BBT, and shallow \abbr). \textit{(2) Modular Prompt vs. Single Prompt}: Shallow \abbr~achieves better performance than SPoT on 13/14 tasks, demonstrating the strong compositional generalization of the modular prompts. \textit{(3) MRC vs. MCC}: PPT lags far behind \abbr~(and even p-tuning v2) on two MRC tasks, namely CMRC-2018 and DRCD, demonstrating the limitation of the MCC format. \textit{(4) Pre-trained Prompt Tuning vs. Prompt Tuning From Scratch}: Pre-trained prompt tuning (i.e., PPT, SPoT, and \abbr) performs consistently better than tuning randomly initialized prompt with the same number of tunable parameters. \textit{(5) Gradient Descent vs. Black-Box Tuning}: Without \abbr~for initialization, BBT and BBTv2 achieve better performance than prompt tuning and p-tuning v2, respectively, on most tasks but much worse performance on a few tasks such as CCPM. By using \abbr~for initialization, the gap between gradient descent and black-box tuning on these tasks are closed, and in average, BBT and BBTv2 outperform their gradient-based counterparts, showing the superiority of gradient-free optimization in few-shot learning settings.

\begin{table}[t]
\resizebox{\linewidth}{!}{
\begin{tabular}{lcccc}
\toprule
\multirow{2}{*}{\textbf{Stage}} & \multicolumn{2}{c}{\textsc{Unseen Data}} & \multicolumn{2}{c}{\textsc{Unseen Task}} \\ \cmidrule(l{5pt}r{8pt}){2-3} \cmidrule(l{5pt}r{5pt}){4-5}
                                & THUCNews             & BQ                & TNews               & LCQMC              \\ \midrule
\multicolumn{5}{c}{Shallow \abbr~with Black-Box Tuning}                                                                   \\ \midrule
Only Stage I                   & 1.26           & 1.10        & 1.61          & 1.11        \\
Two-Stage                      & 14.46          & 7.74        & 25.20         & 6.70         \\ \midrule
\multicolumn{5}{c}{Deep \abbr~with Black-Box Tuning}                                                                      \\ \midrule
Only Stage I                   & 2.62           & 2.90        & 8.20          & 2.28        \\
Two-Stage                      & 7.88           & 5.57        & 17.44         & 4.51         \\ \bottomrule
\end{tabular}
}
\caption{Comparison of training time (in minutes) between two tuning stages.}
\label{tab:two_stage}
\end{table}

\paragraph{Two-Stage Tuning.}
As demonstrated in Table~\ref{tab:main_results}, by only tuning the router (only stage I), which contains merely $8$ parameters for shallow \abbr~or $8\times24=192$ parameters for deep \abbr, we can achieve surprisingly strong performance that can be comparable to two-stage tuning.
For shallow \abbr, only tuning the router even outperforms two-stage tuning in average on \textsc{Unseen Data} tasks.
To take a closer look, we demonstrate the process of two-stage tuning with shallow \abbr~for initialization in Figure~\ref{fig:two_stage}. For both learning paradigms, we find that the best performance on the development set of the \textsc{Unseen Data} task (here is the BQ task) can be observed in stage I, where we only tune the router to reuse pre-trained prompts. On \textsc{Unseen Task} (here is the LCQMC task), we observe improvement of performance during stage II.
In Table~\ref{tab:two_stage}, we compare the training time of the two stages to show the high efficiency of stage I when using black-box tuning. Results suggest that learning to combine instead of tuning the prompts is a promising way to achieve fast adaptation to downstream tasks.

\begin{figure}[t]
    \centering
    \begin{subfigure}{.49\linewidth}
    \centering
    \includegraphics[width=\linewidth]{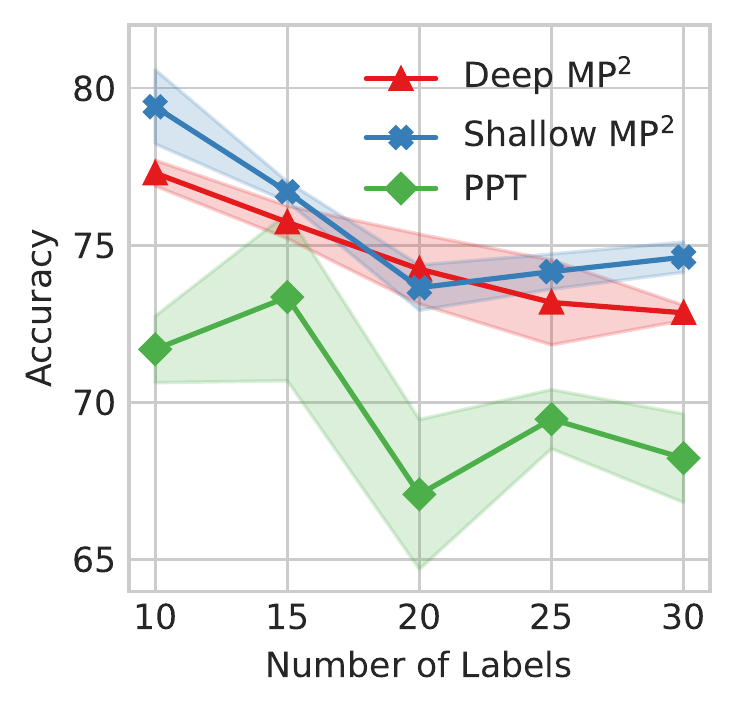}
    \caption{On number of labels}
    \end{subfigure}
    \begin{subfigure}{.49\linewidth}
    \centering
    \includegraphics[width=.95\linewidth]{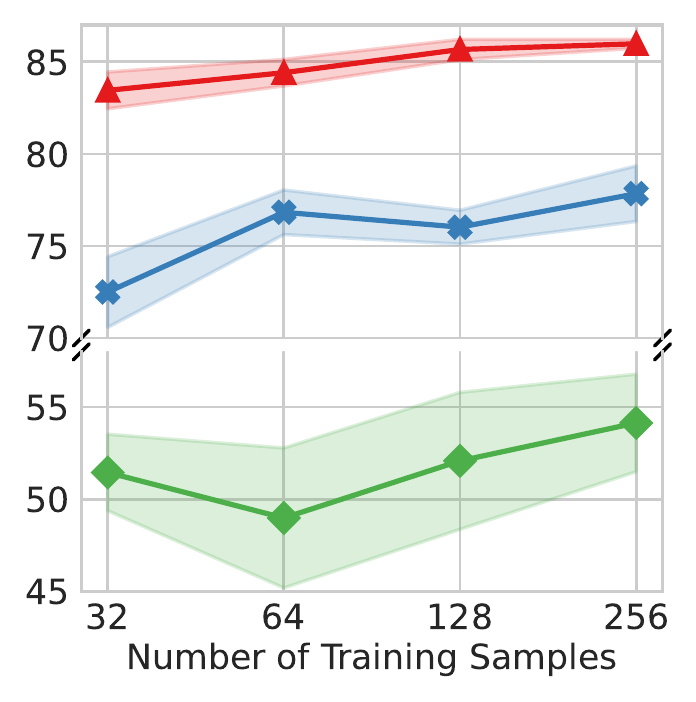}
    \caption{On sample efficiency}
    \end{subfigure}
    \caption{Comparison of \abbr~and PPT with varying numbers of labels and training samples.}
    \label{fig:ablation}
\end{figure}

\paragraph{On Many-Label Classification Tasks.}
In contrast to PPT that is pre-trained to perform up to 16-label classification, our proposed \abbr~unifies tasks into the MRC format such that it can generalize to downstream tasks with varying numbers of labels. To simulate tasks with different numbers of labels, we extract subsets with 10/15/20/25/30 labels from the IFLYTEK dataset, which contains 119 labels in total. We follow the same procedure (§\ref{sec:data_task}) to generate train/dev/test splits from the extracted subsets.
As shown in Figure~\ref{fig:ablation}(a), there is a sharp decline in the accuracy of PPT when the number of labels exceeds 16. By contrast, the performance of \abbr~is decreasing more slowly and steadily as the number of labels increases, demonstrating the superiority of the MRC format.

\paragraph{On Sample Efficiency.}
We compare \abbr~and PPT with different numbers of training samples on the LCQMC task. As shown in Figure~\ref{fig:ablation}(b), increasing training samples generally confers improved performance for both \abbr~and PPT while \abbr~consistently outperforms PPT under varying numbers of training samples. In addition, the gap between \abbr~and PPT cannot be easily filled with enlarged training set.

\paragraph{Task Partitions Induced From the Router.}

\begin{figure}[t]
    \centering
    \includegraphics[width=\linewidth]{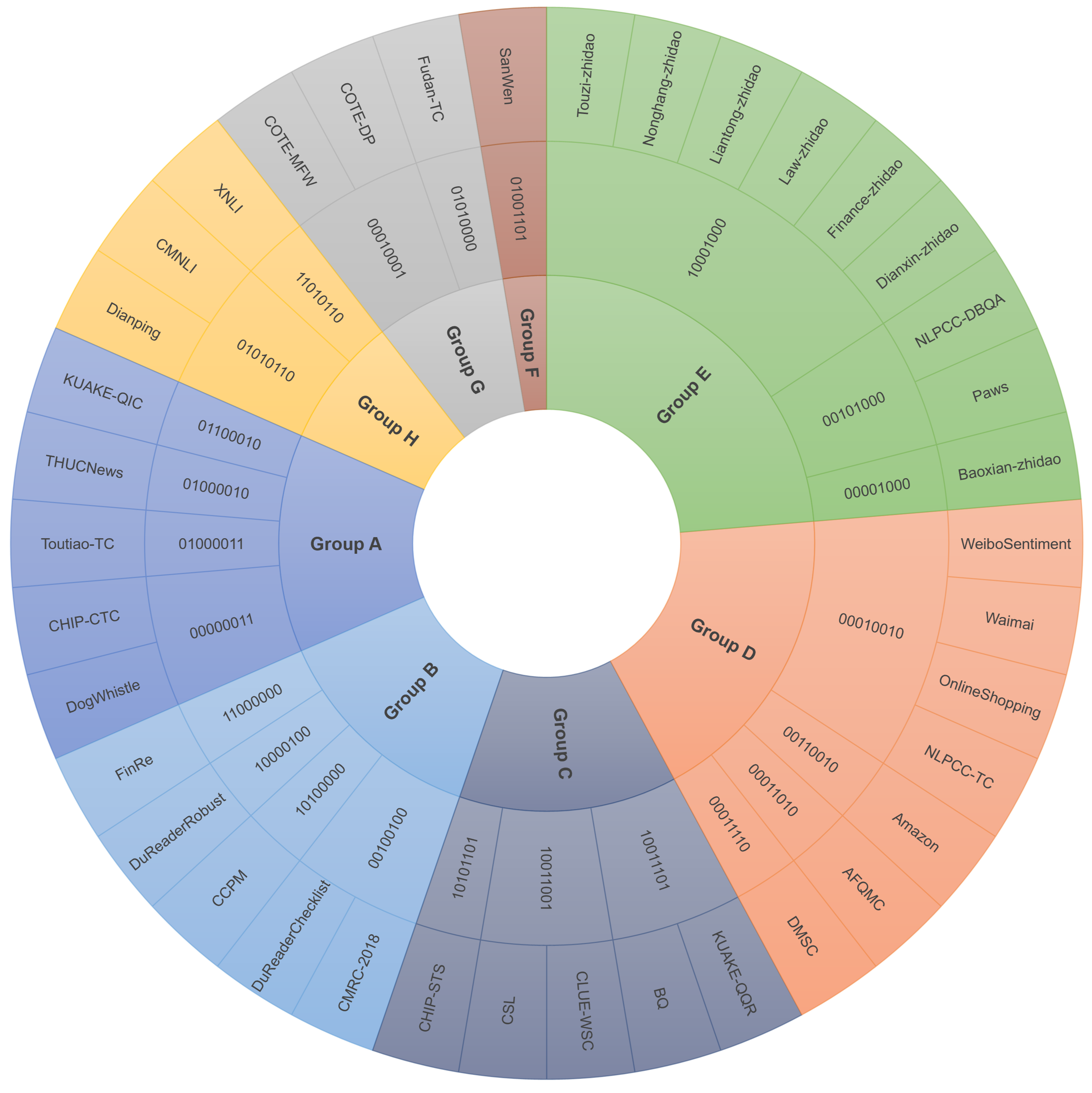}
    \caption{Task partitions induced from the router. Similar tasks are assigned similar subsets of prompts.}
    \label{fig:task_partition}
\end{figure}

We take a closer look at the learned router and find that non-trivial task partitions can be induced from it. For simplicity, we focus on the shallow \abbr, which has only one router. There are totally 8 modular prompts corresponding to $2^8=256$ possible combinations. We perform a hierarchical clustering on the router learned on 38 upstream tasks and visualize the task partitions in Figure~\ref{fig:task_partition}. The 38 upstream tasks can be partitioned into 8 groups. For instance, group A is mainly comprised of topic classification tasks; group D contains all the sentiment analysis tasks; group C and E are all comprised of NLI tasks, among which group E covers all the "Zhidao" tasks, which are question-answer matching tasks. 

\section{Conclusion}
This work aims to bridge the gap between pre-training and fine-tuning of soft prompt tuning for few-shot learning. To achieve this, we extend the soft prompt in two dimensions, depth and width. The extended prompt, named deep modular prompt, is pre-trained on a mixture of 38 public Chinese NLP tasks, which are reformulated into the MRC format. For adaptation to downstream tasks, we propose the two-stage tuning, where we first learn to combine and reuse pre-trained prompts and then tune the selected prompts with gradient descent or black-box optimization. Extensive experiments on 14 downstream tasks demonstrate that, the \textbf{M}ulti-task \textbf{P}re-trained \textbf{M}odular \textbf{P}rompt (\textbf{\abbr}) significantly outperforms prompt tuning, full model tuning, and previous prompt pre-training methods, namely PPT and SPoT. Surprisingly, we demonstrate that \abbr~can achieve extremely fast adaptation to downstream tasks by only learning to combine pre-trained prompts.

\section*{Limitations}
In this work, we demonstrate the effectiveness of the proposed \abbr~with the backbone PTM of CPT-large on a set of Chinese NLP tasks. Due to the expensive pre-training cost, we did not explore \abbr~on other PTMs with varying sizes, pre-training objectives and architectures. Besides, it is also unknown how does the number of pre-training tasks affect the performance of \abbr. For resource-rich languages such as English and Chinese, it would be promising for \abbr~to be well-performed since one can easily collect sufficient public upstream tasks. Nevertheless, for low-resource languages or domains, the effect of \abbr~is still under-explored.

\section*{Ethics Statement}
The proposed \abbr~is a parameter-efficient approach for few-shot learning. In addition, we demonstrate that \abbr~can achieve highly efficient adaptation to a target task by only tuning a few parameters. Therefore, this work helps reduce computation costs and carbon emissions, and can facilitate the adaptation of PTMs to low-resource downstream tasks. Though all the datasets used in our experiments are publicly available and have not been reported to carry social bias against any sensitive attributes, and the proposed approach would not explicitly introduce new negative societal impacts, more work is still needed to investigate the potential unfairness in these datasets.


\section*{Acknowledgements}
This work was supported by the National Natural Science Foundation of China (No. 62236004 and No. 62022027) and CAAI-Huawei MindSpore Open Fund.

\bibliography{anthology,custom}
\bibliographystyle{acl_natbib}

\newpage
\appendix
\section{Implementation Details}
\label{sec:append_implement}
\begin{figure*}
    \centering
    \begin{subfigure}{.8\linewidth}
    \centering
    \includegraphics[width=\linewidth]{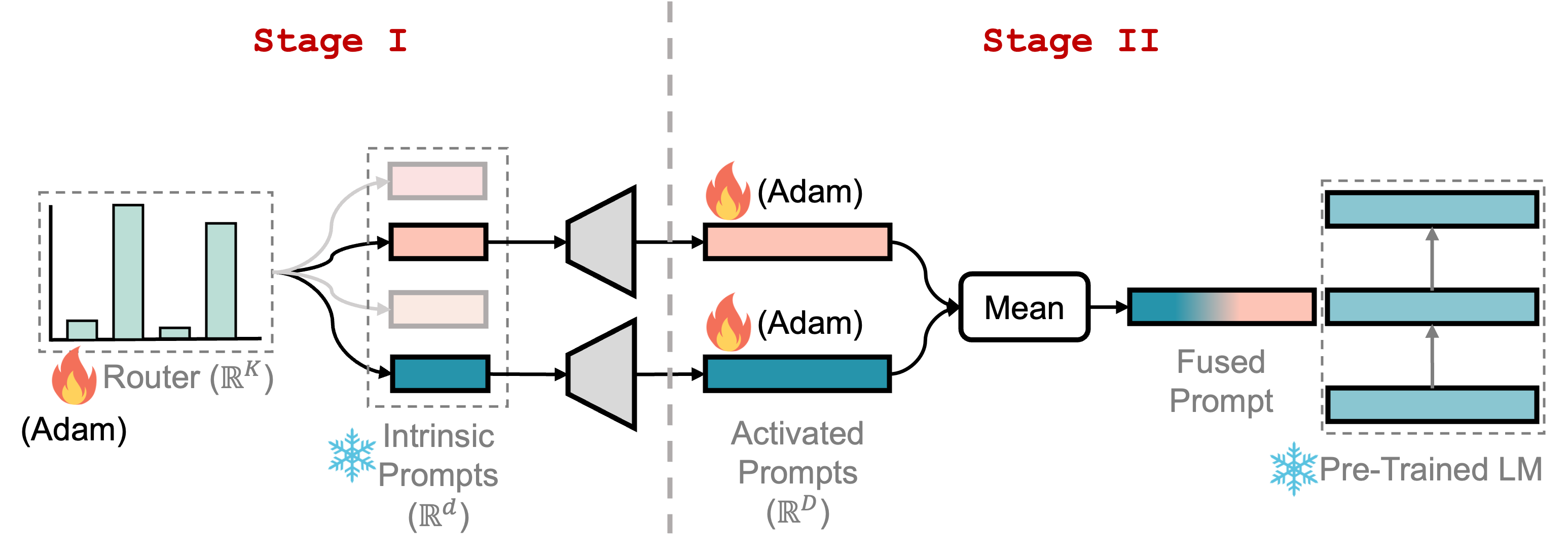}
    \caption{Gradient descent (tunable parameters: $[K+D, K+KD]$ per layer)}
    \end{subfigure}
    \begin{subfigure}{.8\linewidth}
    \centering
    \includegraphics[width=\linewidth]{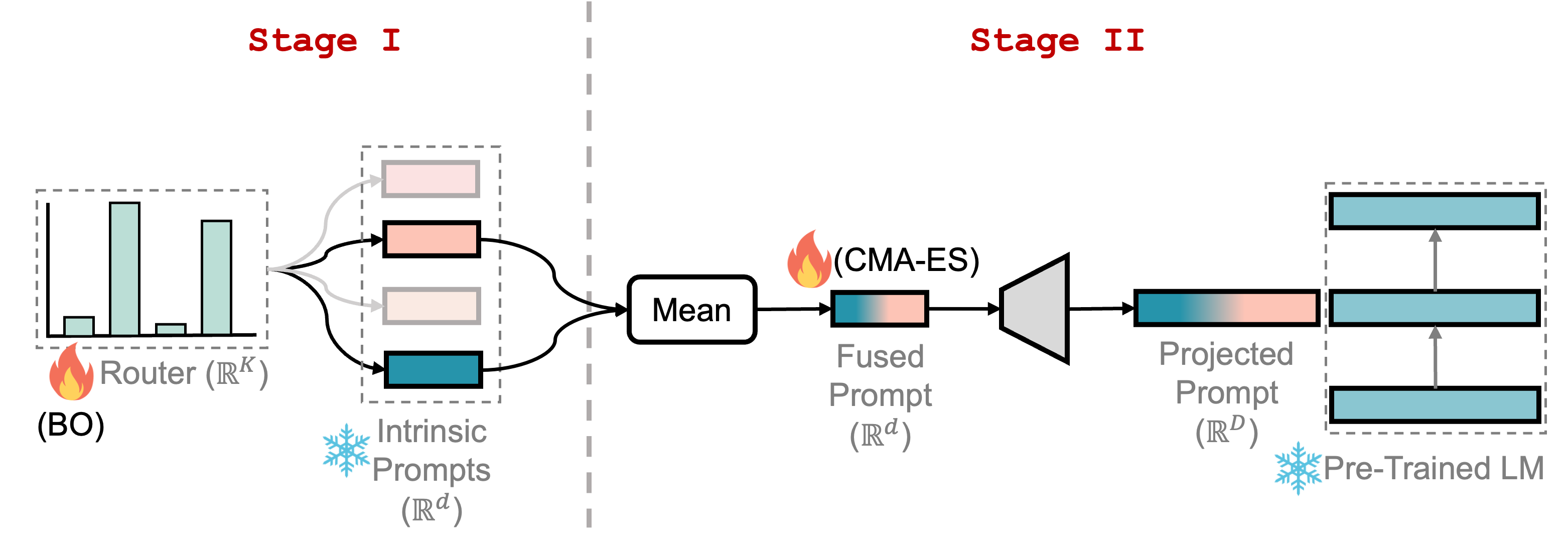}
    \caption{Black-box tuning (tunable parameters: $K+d$ per layer)}
    \end{subfigure}
    \caption{Illustration of the two-stage tuning for gradent descent and black-box tuning. For black-box tuning, which is a gradient-free optimization approach that cannot well handle high-dimensional optimization, we perform pre-fusion to obtain a low-dimensional fused prompt for optimization.}
    \label{fig:two_stage_downstream}
\end{figure*}

\subsection{Upstream Pre-training}
\paragraph{\abbr.}
\abbr~is pre-trained on 38 upstream tasks using an Adam optimizer with batch size of 32 for 1M steps. During each forward computation, we first randomly select a task and then fetch a batch of training data corresponding to the selected task. By this, the number of learning steps on each task is expected to be identical. As demonstrated in Table~\ref{tab:ablation_fsl}, the fast and slow learning (FSL) can be beneficial to deep \abbr, and therefore we use two-speed learning rate for pre-training the routers and the prompts of deep \abbr. In particular, the learning rate of the routers is 5e-4, and the learning rate of the prompts is 1e-4. For shallow \abbr, we use a single learning rate of 1e-3 for the router and the modular prompts. The prompt length is set to 50 for both shallow \abbr~and deep \abbr. For shallow \abbr~and each layer of the deep \abbr~, we allocate $K=8$ modular prompts and one router to combine them. 
In addition to the routers and the prompts, we also train the randomly initialized MRC head on the top of the PTM. The original parameters of the PTM are frozen during pre-training.
We run pre-training on NVIDIA A100 GPUs.

\paragraph{Baselines.}
For fair comparison, we also reimplement PPT and SPoT with the same backbone model as \abbr, i.e., CPT-large. For pre-training PPT, we implement the "Unified PPT" variant, which is to formulate tasks into a unified MCC format, to support a variety of downstream tasks. We follow the experimental setup in the original paper and use 10GB data sampled from the WuDaoCorpora for pre-training. We train for 400K steps using an Adam optimizer with batch size of 32 and learning rate of 3e-2. For SPoT, we pre-trained a single soft prompt on the same 38 upstream tasks as used by \abbr~using an Adam optimizer with batch size of 32 and learning rate of 3e-3 for 650K steps. 
Though the numbers of training steps for PPT and SPoT are less than \abbr, they are sufficient for convergence due to their limited numbers of parameters.
To be consistent with \abbr, we set prompt length to 50 for PPT and SPoT.

\subsection{Downstream Fine-tuning}
We use the two-stage tuning to adapt \abbr~to various downstream tasks. 
In stage I, we only tune the router(s)\footnote{A single router for shallow \abbr~and 24 routers for deep \abbr.} while keeping all other parameters frozen. In stage II, we fix the learned router(s) and only fine-tune the modular prompts selected by the router(s). 
The implementation details of the two-stage tuning can be different for gradient descent and black-box tuning.
We provide a graphical illustration of the two-stage tuning using gradient descent and black-box tuning in Figure~\ref{fig:two_stage_downstream}.
\textbf{For gradient descent}, we fine-tune \abbr~for 1K epochs on each task, where the first 500 epochs as stage I and the last 500 epochs as stage II. For the shallow/deep \abbr, we use an Adam optimizer with learning rate of 1e-2/3e-3 for tuning the router(s) (stage I) and learning rate of 3e-4/2e-5 for tuning the prompts (stage II). 
\textbf{For black-box tuning}, we fine-tune shallow/deep \abbr~for 8K iterations (model forward computes) on each task, where the first 200/100 iterations as stage I and the rest as stage II. In stage I, we use Bayesian optimization (BO) with the acquisition function of upper confidence bound (UCB) with $\kappa=2$ to tune the parameters of the router(s). In stage II, we use CMA-ES to optimize the prompts.
For shallow \abbr, we use $\mu=0$ and $\sigma=0.1$ for initialization of the CMA-ES.
For deep \abbr, we follow BBTv2 and use the divide-and-conquer algorithm to alternately optimize the prompt at each layer. For optimization of the prompt at the embedding layer, we initialize CMA-ES with $\mu=0$ and $\sigma=$~5e-2. For optimization of the prompt at intermediate layers, we adopt $\mu=0$ and $\sigma=$~1e-2. All the hyper-parameters are tuned manually in a lightweight manner on development sets. We perform fine-tuning a single NVIDIA 3090 GPU.

\begin{table}[t]
\centering
\resizebox{\linewidth}{!}{
\begin{tabular}{lcccc}
\toprule
\textbf{Methods} & \textbf{ChnSent} & \textbf{TNews} & \textbf{LCQMC} & \textbf{DRCD} \\ \midrule
\multicolumn{5}{c}{Shallow \abbr}                                                       \\ \midrule
w/o FSL          & 90.46 \textsubscript{0.16}      & 51.36 \textsubscript{1.12}    & 72.50 \textsubscript{1.92}     & 67.20 \textsubscript{2.96}    \\
w/ FSL           & 89.36 \textsubscript{0.63}      & 51.36 \textsubscript{1.30}    & 70.42 \textsubscript{1.27}    & 58.96 \textsubscript{0.73}   \\ \midrule
\multicolumn{5}{c}{Deep \abbr}                                                          \\ \midrule
w/o FSL          & 91.61 \textsubscript{0.18}      & 55.23 \textsubscript{0.29}    & 82.30 \textsubscript{1.28}    & 78.69 \textsubscript{0.72}   \\
w/ FSL           & 92.02 \textsubscript{0.11}      & 54.71 \textsubscript{0.31}    & 83.45 \textsubscript{1.00}     & 80.64 \textsubscript{0.87}   \\ \bottomrule
\end{tabular}
}
\caption{Ablation of fast and slow learning (FSL).}
\label{tab:ablation_fsl}
\end{table}

\begin{table}[t]
    \centering
    \resizebox{\linewidth}{!}{
    \begin{tabular}{ll}
    \toprule
        \textbf{Dataset} & \textbf{Source}\\ \midrule
        ChnSent & \url{https://github.com/SophonPlus/ChineseNlpCorpus}\\
        TNews & \url{https://github.com/fatecbf/toutiao-text-classfication-dataset/}\\
        OCNLI & \citet{Hu2020OCNLI}\\
        LCQMC & \citet{Liu2018LCQMC}\\
        DRCD & \citet{Shao2018DRCD}\\
        C$^3$ & \citet{Sun2020C3}\\
        COTE-BD & \citet{li2018cote}\\
         \bottomrule
    \end{tabular}
    }
    \caption{Sources of the 7 downstream tasks in the \textsc{Unseen Task} track.}
    \label{tab:source_unseen_task}
\end{table}

\section{Additional Results}
\label{sec:append_add_res}

\paragraph{Ablation of Fast and Slow Learning.}
We conduce ablation study on fast and slow learning (FSL), which is to assign different learning rates to routers and prompts. As demonstrated in Table~\ref{tab:ablation_fsl}, FSL exhibits positive effect on downstream tasks to deep \abbr~and negative effect to shallow \abbr. Therefore, we retain the shallow \abbr~pre-trained without FSL and the deep \abbr~pre-trained with FSL in our experiments.

\section{MRC Format}
\label{sec:append_format}
We unify upstream and downstream tasks into the machine reading comprehension (MRC) format, which takes as input a \textit{context} and a \textit{query}, and outputs the \textit{answer} of the query. For topic classification and sentence-pair classification tasks, we use the original input text as the context and construct a query containing all valid labels. The context and the constructed query are concatenated and fed into the model. The model is trained to extract the answer in the query by predicting its start and end positions. For more complicated tasks such as relation extraction and poem understanding, we manually design task-specific templates to convert inputs to the desired contexts and queries. Some examples are shown in Table~\ref{tab:MRC_template}.

\begin{table*}[t]
\resizebox{\linewidth}{!}{
\begin{tabular}{lcl}
\toprule
\textbf{Dataset}        & \textbf{Task}         & \textbf{Template}                                                                                                                                                                                          \\ \midrule
\multirow{2}{*}{Amazon} & \multirow{2}{*}{TC}   & \begin{CJK}{UTF8}{gbsn}打分：$\langle S\rangle$的评价是？选项：非常差，较差，一般，较好，非常好。\end{CJK} \\
& & (Rating: $\langle S\rangle$ What do you think about it? Options: very bad, bad, okay, good, very good.)\\ \midrule
\multirow{2}{*}{ChnSent} & \multirow{2}{*}{TC}   & \begin{CJK}{UTF8}{gbsn}情感分析：$\langle S\rangle$的情感是？选项：负面，正面。\end{CJK} \\
& & (Sentiment analysis: What is the sentiment of $\langle S\rangle$? Options: negative, positive.)\\ \midrule
\multirow{2}{*}{TNews}  & \multirow{2}{*}{TC}   & \begin{CJK}{UTF8}{gbsn}主题识别：$\langle S\rangle$的主题是？选项：房产，汽车，金融，体育，文化...\end{CJK}\\
& & (Topic classification: What is the topic of $\langle S\rangle$? Options: housing, car, finance, sports, culture, ...)\\\midrule
\multirow{2}{*}{FinRe}  & \multirow{2}{*}{TC}   & \begin{CJK}{UTF8}{gbsn}关系判别：$\langle S1\rangle$和$\langle S2\rangle$在句子中的关系是？选项：未知，注资，拥有，纠纷，自己...\end{CJK}\\
& & (Relation classification: What is the relationship between $\langle S1\rangle$ and $\langle S2\rangle$? Options: unknown, capital injection, possess, dispute, oneself...)\\\midrule
\multirow{2}{*}{CMNLI}  & \multirow{2}{*}{NLI}  & \begin{CJK}{UTF8}{gbsn}意思判别：$\langle S1\rangle$与$\langle S2\rangle$的关系是？选项：矛盾，蕴含，中立。\end{CJK}\\
& & (Textual entailment: What is the relationship between $\langle S1\rangle$ and $\langle S2\rangle$? Options: contradiction, entailment, neutral.)\\\midrule
\multirow{2}{*}{CCPM}   & \multirow{2}{*}{MCQA} & \begin{CJK}{UTF8}{gbsn}诗句理解：与句子$\langle S\rangle$最相近的诗句是？选项：$\langle A1\rangle$，$\langle A2\rangle$，$\langle A3\rangle$，$\langle A4\rangle$。\end{CJK}\\
& & (Poem understanding: Which verse comes closest to $\langle S\rangle$? Options: $\langle A1\rangle$, $\langle A2\rangle$, $\langle A3\rangle$, $\langle A4\rangle$.) \\\midrule
\multirow{2}{*}{C$^3$}     & \multirow{2}{*}{MCQA} & \begin{CJK}{UTF8}{gbsn}阅读选择：文档：$\langle S1\rangle$，问题：$\langle S2\rangle$，选项：$\langle A1\rangle$，$\langle A2\rangle$，$\langle A3\rangle$。\end{CJK}\\
& & (Reading comprehension: Document: $\langle S1\rangle$, Question: $\langle S2\rangle$, Options: $\langle A1\rangle$, $\langle A2\rangle$, $\langle A3\rangle$.) \\ \bottomrule
\end{tabular}
}
\caption{Example templates to formulate non-MRC tasks into the MRC format.}
\label{tab:MRC_template}
\end{table*}

\begin{table*}[t]
\resizebox{\linewidth}{!}{
\begin{tabular}{llccccccl}
\toprule
\textbf{ID} & \textbf{Dataset} & \textbf{Task} & \textbf{Domain} & \textbf{|Train|} & \textbf{|Dev|} & \textbf{|Test|} & \textbf{|Labels|} & \textbf{Reference} \\ \midrule
1 & AFQMC & NLI & Financial & 31k & 3k & 4k & 2 & \citet{Xu2020CLUE} \\
2 & Paws & NLI & General & 44k & 5k & 2k & 2 & \citet{Yang2019paws} \\
3 & CMNLI & NLI & General & 380k & 12k & 12k & 3 & \citet{Xu2020CLUE} \\
4 & CSL & NLI & Academic & 18k & 2k & 3k & 2 & \citet{Xu2020CLUE} \\
5 & BQ & NLI & Financial & 90k & 10k & 10k & 2 & \citet{Chen2018BQ} \\
6 & CHIP-STS & NLI & Biomedical & 14k & 2k & 4k & 2 & \citet{Zhang2018CBLUE} \\
7 & KUAKE-QQR & NLI & Clinical & 14k & 2k & 2k & 3 & \citet{Zhang2018CBLUE} \\
8 & XNLI & NLI & General & 380k & 12k & 2k & 3 & \citet{Conneau2018Xnli} \\
9 & NLPCC-DBQA & NLI & General & 170k & 12k & 41k & 2 & \url{http://tcci.ccf.org.cn/conference/2016} \\
10 & Finance-zhidao & NLI & Financial & 64k & 12k & 38k & 2 & \url{https://github.com/SophonPlus/ChineseNlpCorpus} \\
11 & Law-zhidao & NLI & Law & 23k & 3k & 7k & 2 & \url{https://github.com/SophonPlus/ChineseNlpCorpus} \\
12 & Liantong-zhidao & NLI & Telecom & 150k & 12k & 20k & 2 & \url{https://github.com/SophonPlus/ChineseNlpCorpus} \\
13 & Nonghang-zhidao & NLI & Financial & 29k & 3k & 4k & 2 & \url{https://github.com/SophonPlus/ChineseNlpCorpus} \\
14 & Touzi-zhidao & NLI & Investment & 487k & 12k & 29k & 2 & \url{https://github.com/SophonPlus/ChineseNlpCorpus} \\
15 & Baoxian-zhidao & NLI & Insurance & 5k & 0.6k & 2k & 2 & \url{https://github.com/SophonPlus/ChineseNlpCorpus} \\
16 & Dianxin-zhidao & NLI & Telecom & 99k & 11k & 31k & 2 &  \url{https://github.com/SophonPlus/ChineseNlpCorpus} \\
17 & THUCNews & TC & General & 45k & 5k & 5k & 10 & \url{https://github.com/thunlp/THUCTC} \\
18 & CHIP-CTC & TC & Biomedical & 23k & 8k & 10k & 44 & \citet{Zong2021CHIPCTC}   \\
19 & FinRe & TC & Financial & 12k & 1k & 1k & 44 & \citet{Li19Finre}\\
20 & Fudan-TC & TC & General & 9k & 1k & 10k & 20 & Not found$^\dag$ \\
21 & KUAKE-QIC & TC & Clinical & 6k & 0.7k & 2k & 11 & \citet{Zhang2018CBLUE} \\
22 & NLPCC-TC & TC & General & 6k & 0.7k & 2k & 2 & \url{http://tcci.ccf.org.cn/conference/2016} \\
23 & Amazon & TC & Shopping review & 3.6M & 12k & 41k & 5 & \url{https://github.com/SophonPlus/ChineseNlpCorpus} \\
24 & DianPing & TC & Shopping review & 2.6M & 12k & 30k & 5 & \url{https://github.com/SophonPlus/ChineseNlpCorpus} \\
25 & DMSC & TC & Movie review & 1.6M & 12k & 92k & 5 & \url{https://github.com/SophonPlus/ChineseNlpCorpus} \\
26 & Online-Shopping & TC & Shopping review & 45k & 5k & 6k & 2 & \url{https://github.com/SophonPlus/ChineseNlpCorpus} \\
27 & Waimai & TC & Shopping review & 8k & 0.8k & 2k & 2 & \url{https://github.com/SophonPlus/ChineseNlpCorpus} \\
28 & Weibo-sentiment & TC & General & 76k & 8k & 24k & 5 & \url{https://github.com/SophonPlus/ChineseNlpCorpus} \\
29 & Toutiao-TC & TC & General & 321k & 12k & 11k & 14 & \url{https://github.com/aceimnorstuvwxz/toutiao-text-classfication-dataset}\\
30 & SanWen & TC & Literature & 13k & 1k & 2k & 10 & \citet{Xu17Sanwen}\\
31 & CLUE-WSC & CR & General & 1k & 0.1k & 0.3k & 2 & \citet{Xu2020CLUE} \\
32 & COTE-DP & KE & Shopping review & 16k & 2k & 5k & N/A & \citet{li2018cote} \\
33 & COTE-MFW & KE & Shopping review & 26k & 3k & 8k & N/A & \citet{li2018cote} \\
34 & DuReader-Checklist & MRC & General & 3k & 0.3k & 1k & N/A & \url{https://github.com/baidu/DuReader} \\
35 & DuReader-Robust & MRC & General & 13k & 1k & 1k & N/A & \citet{Tang21Dureaderrobust} \\
36 & CMRC-2018 & MRC & General & 8k & 0.9k & 3k & N/A & \citet{Xu2020CLUE} \\
37 & CCPM & MCQA & Poem & 19k & 2k & 3k & 4 & \url{https://github.com/SophonPlus/ChineseNlpCorpus} \\
38 & DogWhistle & MCQA & General & 218k & 12k & 29k & 4 & \citet{xu2021dogwhistle} \\
\midrule
 & Total & - & - & 10.7M & 213k & 499k & - & - \\
\bottomrule
\end{tabular}
}
\caption{Pre-training datasets for \abbr. NLI: natural language inference. TC: text classification. CR: coreference resolution. KE: keyword extraction. MRC: machine reading comprehension. MCQA: multiple choice question answering. $^\dag$ We did not find the official source of the Fudan-TC dataset.}
\label{tab:pretrain_task}
\end{table*}

\section{Additional Details of Tasks}
\label{sec:append_pretrain_task}

\subsection{Upstream Tasks}
Table~\ref{tab:pretrain_task} contains details of the 38 upstream tasks. We only use the training sets during pre-training. For tasks that also serve as a downstream task in the \textsc{Unseen Data} track, we remove a small portion of training samples from pre-training to avoid data leakage.

\subsection{Downstream Tasks}
The downstream tasks are divided into two tracks, \textsc{Unseen Data} and \textsc{Unseen Task}. The tasks in the \textsc{Unseen Data} track are a subset of upstream task, for which the details have been provided in Table~\ref{tab:pretrain_task}. For the 7 tasks in the \textsc{Unseen Task} track, we provide the sources in Table~\ref{tab:source_unseen_task}.

\end{document}